\documentclass[sigconf]{acmart}
\AtBeginDocument{%
  }

\copyrightyear{2026}
\acmYear{2026}
\setcopyright{cc}
\setcctype{by}
\acmConference[FDG '26]{Foundations of Digital Games}{August 10--13, 2026}{Copenhagen, Denmark}
\acmBooktitle{Foundations of Digital Games (FDG '26), August 10--13, 2026, Copenhagen, Denmark}
\acmDOI{10.1145/3815598.3815627}
\acmISBN{979-8-4007-2495-4/2026/08}

\usepackage{subfigure}
\usepackage[acronym]{glossaries}
\usepackage[retain-zero-uncertainty=true,separate-uncertainty=true,text-series-to-math=true,
propagate-math-font=true,detect-all=true]{siunitx}

\newcommand{\minisection}[1]{\noindent {\bf #1} \ }

\makeglossaries

\newacronym{ai}{AI}{Artificial Intelligence}
\newacronym{ssir}{SSIR}{Self-supervised Intrinsic Reward}
\newacronym{ml}{ML}{Machine Learning}
\newacronym{nlp}{NLP}{Natural Language Processing}
\newacronym{api}{API}{Application Programming Interface}
\newacronym{marl}{MARL}{Multi-Agent Reinforcement Learning}
\newacronym{rl}{RL}{Reinforcement Learning}
\newacronym{drl}{Deep RL}{Deep Reinforcement Learning}
\newacronym{dqn}{DQN}{Deep Q-Network}
\newacronym{rnd}{RND}{Random Network Distillation}
\newacronym{grf}{GRF}{Google Research Football}
\newacronym{ppo}{PPO}{Proximal Policy Optimization}
\newacronym{mat}{MAT}{Multi-Agent Transformer}
\newacronym{sp}{SP}{Self-Play}
\newacronym{ne}{NE}{Nash equilibrium}
\newacronym{psro}{PSRO}{Policy-Space Response Oracle}
\newacronym{ddp}{DDP}{DistributedDataParallel}
\newacronym{ctde}{CTDE}{Centralized Training with Decentralized Execution}
\newacronym{jrpo}{JRPO}{Joint-ratio Policy Optimization}
\newacronym{mlp}{MLP}{Multi-layer Perceptron}
\newacronym{lstm}{LSTM}{Long Short-term Memory}
\newacronym{relu}{ReLU}{Rectified Linear Unit}
\newacronym{mse}{MSE}{Mean Squared Error}
\newacronym{gae}{GAE}{Generalized Advantage Estimation}
\newacronym{a2c}{A2C}{Advantage Actor-Critic}
\newacronym{a3c}{A3C}{Asynchronous Advantage Actor-Critic}
\newacronym{impala}{IMPALA}{Importance Weighted Actor-Learner Architecture}
\newacronym{sac}{SAC}{Soft Actor-Critic}
\newacronym{acer}{ACER}{Actor-Critic With Experience Replay}
\newacronym{vdn}{VDN}{Value Decomposition Network}
\newacronym{maddpg}{MADDPG}{Multi-Agent Deep Deterministic Policy Gradient}
\newacronym{ippo}{IPPO}{Independent PPO}
\newacronym{mappo}{MAPPO}{Multi-Agent PPO}
\newacronym{fxp}{FXP}{Fictitious Cross-Play}
\newacronym{pbrs}{PBRS}{Potential-based Reward Shaping}
\newacronym{im}{IM}{Intrinsic Motivation}
\newacronym{mcts}{MCTS}{Monte Carlo Tree Search}
\newacronym{ann}{ANN}{Artificial Neural Network}
\newacronym{mdp}{MDP}{Markov Decision Process}
\newacronym{pomdp}{POMDP}{Partially Observable Markov Decision Process}
\newacronym{ale}{ALE}{Arcade Learning Environment}
\newacronym{cnn}{CNN}{Convolutional Neural Network}
\newacronym{pbt}{PBT}{Population-based Training}
\newacronym{mpe}{MPE}{Multi-agent Particle Environment}
\newacronym{smac}{SMAC}{StarCraft Multi-agent Challenge}
\newacronym{td}{TD}{Temporal Difference}
\newacronym{ucb}{UCB}{Upper Confidence Bound}
\newacronym{solir}{SOLIR}{Self-supervised Online Learned Intrinsic Reward}
\newacronym{hai}{HAI}{heuristic AI}
\newacronym{npc}{NPC}{Non-Player Character}

\glsdisablehyper

\begin{document}

\title{Improving Sample Efficiency in Multi-Agent Reinforcement Learning for Simulated Football Games via Exploration}

\author{Amir Baghi}
\orcid{0009-0008-7041-1944}
\affiliation{
    \institution{SEED - Electronic Arts (EA)}
    \city{Stockholm}
    \country{Sweden}
}
\email{abaghi@ea.com}

\author{Jens Sj{\"o}lund}
\orcid{0000-0002-9099-3522}
\affiliation{
    \department{Dept. of Information Technology}
    \institution{Uppsala University}
    \city{Uppsala}
    \country{Sweden}
}
\email{jens.sjolund@it.uu.se}

\author{Joakim Bergdahl}
\orcid{0000-0001-5720-2533}
\affiliation{
    \institution{SEED - Electronic Arts (EA)}
    \city{Stockholm}
    \country{Sweden}
}
\email{jbergdahl@ea.com}

\author{Linus Gissl\'en}
\orcid{0009-0005-8205-8218}
\affiliation{
    \institution{SEED - Electronic Arts (EA)}
    \city{Stockholm}
    \country{Sweden}
}
\email{lgisslen@ea.com}

\author{Alessandro Sestini}
\orcid{0000-0001-5496-5770}
\affiliation{
    \institution{SEED - Electronic Arts (EA)}
    \city{Stockholm}
    \country{Sweden}
}
\email{asestini@ea.com}

\renewcommand{\shortauthors}{Baghi et al.}

\begin{abstract}
    Multi-agent reinforcement learning has shown promise in learning cooperative behaviors in team-based environments. However, such methods often demand extensive training time, which inhibits their application for game-AI in standard game development. For instance, the state-of-the-art method TiZero takes 40 days to train high-quality policies for a football environment. In this paper, we hypothesize that better exploration mechanisms can improve the sample efficiency of multi-agent methods. Thereby, we propose utilizing a random network distillation bonus within the multi-agent TiZero framework, aiming to promote exploration. Additionally, we introduce architectural modifications to the original algorithm to enhance TiZero’s computational efficiency. We evaluate the sample efficiency of our approach against original TiZero through extensive experiments. Our results show that random network distillation improves the sample efficiency per training phase by 13.3\% compared with the original TiZero, enhancing generalization and adaptability to previously difficult scenarios. This highlights the better applicability of our variant in practical game development settings. Lastly, we qualitatively evaluate the gameplay of the produced models against a heuristic AI. We find that random network distillation leads to a higher accuracy in shooting, and it achieves higher behavioral stability as shown by the lower standard deviation achieved in gameplay metrics. The code is available at \textbf{\url{https://github.com/electronicarts/marling}}.
\end{abstract}

\ccsdesc[500]{Computing methodologies~Multi-agent systems}
\ccsdesc[500]{Computing methodologies~Multi-agent reinforcement learning}
\ccsdesc[500]{Applied computing~Computer games}

\keywords{Multi-agent Reinforcement Learning, Sample Efficiency, Exploration, Game-AI}


\maketitle

\section{Introduction} \label{sec:introduction}

\Gls{marl} is a natural extension of \Gls{rl} to scenarios where multiple learning agents need to interact -- through cooperation or competition -- to reach a pre-defined team-based or globally common goal \cite{marl-book}. Its use spans a wide range of use-cases and problem spaces, such as multi-robot warehouse management and autonomous driving of multiple vehicles \cite{Dinneweth2022_marl_autonomous_driving_survey,krnjaic2024multi_robot_warehouse_management}.\looseness=-1 

Recently, \gls{marl} has been employed as game-AI in multi-player and multi-character game environments, where cooperation and competition are common challenges facing human players and bots alike \cite{vinyals2019grandmaster_league_sp,xu2024higher}. A notable example is AlphaStar, which successfully applied \gls{marl} in the real-time strategy video game \textit{StarCraft~II} \cite{vinyals2019grandmaster_league_sp}.
Another use-case for \gls{marl} is controlling a team of football players in a game \cite{kurach2020gfootball,lin2023tizero}. Football is a useful testbed for multi-agent control methods, as it encapsulates the challenges of strategy, cooperation, and competition. Moreover, it has proven difficult to create human-like, emergent and representative team-based behavior for football via hand-crafted and scripted methods, which often fail to adapt dynamically to new situations \cite{lara-cabrera_game_2015}.

Many notable \gls{marl} methods have been proposed that successfully learn useful policies for different domains \cite{xu2024higher,lin2023tizero}. However, many of these methods suffer from poor sample efficiency, ultimately leading to excessive training times for reaching gameplay-suitable policies. For example, in the work by Lin et al. \cite{lin2023tizero}, the authors report a training time of $\sim 40$ days across hundreds of processors. In this work, we recognize this as a significant drawback of such methods which inhibits their tractability in traditional game development settings where rapid iteration is essential. We further hypothesize that the sample efficiency of these methods can be improved via \textit{better exploration}, and thus we tackle the problem from this perspective. Our contribution here is empirical, i.e., we investigate this hypothesis through extensive experiments and ablations that showcase how better exploration translates into sample-efficiency gains, exhibited within the challenging team-based football domain. Additionally, we limit the scope and duration of our experiments here to a few days rather than multiple weeks or long horizons such as 40 days, due to our focus on a more rapid and practical development framework. 

\begin{figure}
    \centering
    \includegraphics[width=0.85\columnwidth]{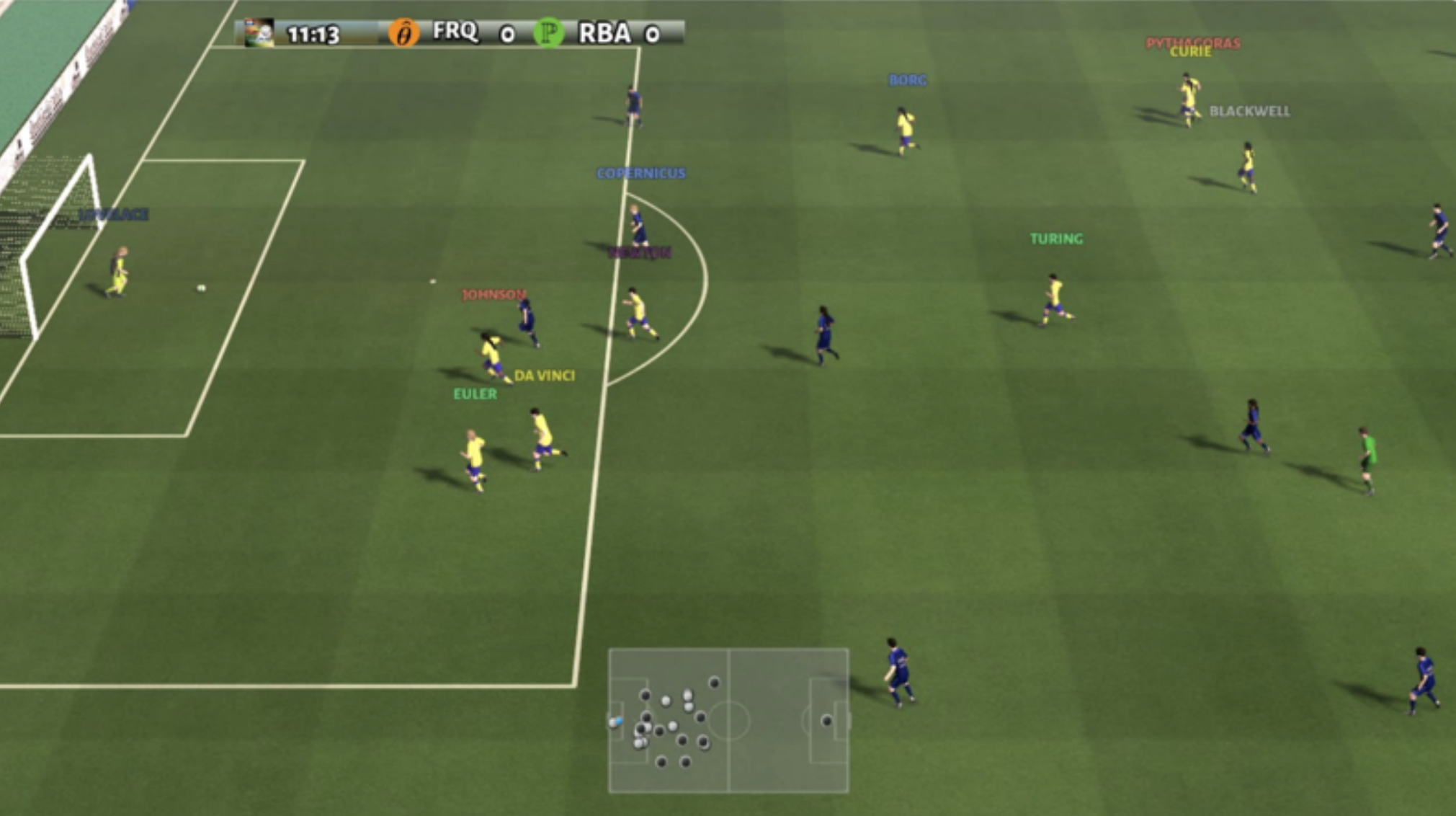} 
    \caption{A screenshot of an 11-vs-11 scenario in the Google research football environment we use as a testbed for our training experiments and evaluations.}
    \label{fig:grf-screenshot}
    \Description{A screenshot of a 11-vs-11 level in Google research football environment.}
\end{figure}

Thus, we focus on improving \textit{TiZero}, a state-of-the-art multi-agent algorithm for team-based football game-AI \cite{lin2023tizero}.
Specifically, besides methodological and architectural changes, we propose a different exploration-promoting reward extension to TiZero: utilizing an exploration bonus in the form of \Gls{rnd} \cite{burda2018rnd}. We apply this technique, originally proposed for single-agent \gls{rl}, to \gls{marl}, and train and evaluate the resulting methods in the football environment provided by \Gls{grf} \cite{kurach2020gfootball} and report both training statistics and qualitative gameplay metrics against a heuristic AI. Fig.~\ref{fig:grf-screenshot} shows a screenshot of the environment.

In summary, our contributions and results are the following:
\begin{enumerate}
    \item We propose methodological and architectural modifications aimed at improving computational performance, as well as an exploration bonus in the form of \gls{rnd}, to enhance the sample efficiency of TiZero. We then empirically demonstrate their effects on training and gameplay performance.
    \item Restricting training to a few days to mirror realistic game-development settings, we evaluate the qualitative behavior of the team-based policies produced by the original TiZero and TiZero augmented with \gls{rnd} in the \gls{grf} environment against a baseline \gls{hai} which was less encountered during training. The results show that each method reaches a gameplay style with its own distinct attributes, with TiZero-RND producing a strategy with a higher accuracy in shooting. It further shows higher stability and certainty in its converged behavior compared to the original TiZero due to lower variance in reported metrics.
    \item We conduct an ablation study of fixed positional encoding of player IDs and its impact on training sample efficiency and stability compared to the original MLP-based encoding approach.
\end{enumerate}

\section{Related Work} \label{sec:related-work}

In this section, we review relevant works in \gls{rl} and \gls{marl} with a special focus on game-AI, football, reward shaping, and \gls{rl} exploration.

\subsection{Reinforcement Learning for Game-AI}
\gls{rl} has been gaining interest in the game AI and game development communities due to its potential to generate human-like and adaptive \acrfull{npc} behaviors. Notable examples, such as AlphaStar, OpenAI 5,  and GT Sophy~\cite{openai2019dota, vinyals2019grandmaster_league_sp, wurman2022outracing}, demonstrate the potential of \gls{rl} to reach and even surpass human-level performance. However, the aforementioned approaches primarily focus on achieving super-human performances, regardless of the computational constraints. For instance, the work by OpenAI~\cite{openai2019dota} required months of training. This significantly reduces the applicability of such approaches in a real-world game development pipeline \cite{jacob2020s}.

At the same time, recent literature has shifted towards developing agents that are practical for use within game production environments. Some notable examples include: the work by \citet{zhang2024training} that trained a DRL agent for the game Arena Breakout; the work by \citet{sestini2025human}, who proposed a sample-efficient single-agent framework enabling game designers to train and modify the agent within a few hours;  and Sh\=ukai, a practical DRL algorithm specifically tailored for commercial fighting games~\citep{zhang2024advancing}. Although these works recognize sample inefficiency as one of the primary challenges limiting the applicability of \gls{rl} in production, they largely focus on single-agent settings. Our study addresses this gap by presenting initial findings on improving sample efficiency in \gls{marl}, towards a more affordable multi-agent game-AI system.

\subsection{Multi-Agent Reinforcement Learning in Football} \label{sec:related-work-marl-football}
The game of football has been proven to be a challenging multi-agent testbed. Therefore, recent research tackles this as a \gls{marl} problem.
As a state-of-the-art method, TiZero employs a curriculum-based self-play strategy, learning a cooperative policy via modifications to the \gls{mappo} policy loss while employing \gls{ctde} \cite{lin2023tizero}. This method reaches a high win-rate of $\sim 95\%$ against strong baselines. Fictitious cross-play -- another state-of-the-art \gls{marl} method -- couples self-play with the policy-space response oracle technique to achieve win-rates exceeding $94\%$ in \gls{grf} \cite{xu2023fxp}. In this paper, we base our work on TiZero as an exemplary and effective \gls{marl} method for football. However, TiZero and similar \gls{marl} methods are known to be sample inefficient, with the former requiring 40 days of training to reach strong gameplay performance. Similar to recent research, such as the work by Xu et al.~\cite{xu2024higher}, we aim to improve sample efficiency through reward shaping and exploration.

Among the available \gls{marl} football environments, \textit{\acrfull{grf}} offers an intermediate level of abstraction for high-level team-based strategies while maintaining individual player control \cite{kurach2020gfootball}. In contrast, JiDi Olympics Football simplifies the game by focusing primarily on movement and ball control without complex team strategies \cite{wang2024jidi_olympics}. At the other extreme, MuJoCo Multi-Agent Soccer emphasizes low-level robotic control of players \cite{deepmind_mujoco}. The balanced nature of \gls{grf} makes it the preferred choice for our work, as it encapsulates both short-term individual skills and long-term strategic decision-making. 

\subsection{Reward Shaping and Exploration}

Reward shaping is a technique in \gls{rl} where the reward function is augmented with an additional term based on the agent transitions and heuristics regarding the desired goal, environment, and general agent behavior. It has been shown to accelerate learning by providing a dense, guiding, and intermediate signal of reinforcement in sparse environments \cite{ng1999policy_invariance_potential_based_rew_shaping}.

A common form of reward shaping is intrinsic motivation. This technique rewards the agent for the inherent ``satisfaction'' of discovering new knowledge and skills \cite{chentanez2004intrinsically_motivated_rl}. Intrinsic motivation thereby encourages agents to autonomously discover skills that can be useful to achieve the end goal.
As an example technique in reward shaping and intrinsic motivation, \acrfull{rnd} encourages exploration by providing intrinsic rewards based on prediction errors on a fixed, randomly-initialized neural network~\cite{burda2018rnd}.\looseness=-1

Specifically within exploration, a commonly employed technique in \gls{rl} is entropy regularization, which promotes an exploratory optimization process by encouraging the agent to maintain stochasticity in its action selection. It has been utilized in the actor loss function of policy gradient methods, e.g., \gls{mappo} \cite{zhang2024entropy_regularized_diffusion_policy,yu2022mappo_ippo}. TiZero, which is the basis of our study, utilizes entropy regularization in its actor loss function as well, and so, it already benefits from one form of exploration.

Beyond entropy regularization, \gls{rnd} has shown effective exploration even in games~\cite{sestini2022automated}, and thus, we employ it in this work to address the poor sample efficiency of the \gls{marl} algorithm under study.

\section{Preliminaries and Setting} \label{sec:prelims}
Here, we summarize relevant theoretical concepts that will help the reader understand subsequent sections. These include the \gls{marl} setting and the training environment.

\subsection{Multi-Agent Reinforcement Learning}

A \gls{marl} problem is typically modeled as a stochastic game, which generalizes the \acrfull{mdp} to multiple agents \cite{littman1994markov_games}. A stochastic game is formally defined as:
\begin{equation}
\mathcal{M} = (\mathcal{N}, \mathcal{S}, \{\mathcal{A}_i\}_{i \in \mathcal{N}}, P, \{r^i\}_{i \in \mathcal{N}}, \gamma),
\end{equation}
where \( \mathcal{N} = \{1, 2, \dots, n\} \) is the set of agents, \( \mathcal{S} \) is the state space, and \( \mathcal{A}_i \) denotes the action space of agent \( i \). The transition function \( P: \mathcal{S} \times \mathcal{A}_1 \times \dots \times \mathcal{A}_n \times \mathcal{S} \to [0,1] \) defines the probability of reaching the next state given the current environment state and joint actions, and each agent receives a reward \( r^i: \mathcal{S} \times \mathcal{A}_1 \times \dots \times \mathcal{A}_n \to \mathbb{R} \). Moreover, the discounting factor is denoted by \( \gamma \in [0,1] \).

At each time step \( t \), the environment is in state \( s_t \), and each agent selects an action \( a_t^i \) according to its policy \( \pi^i(a_t^i | o_t^i) \), where \( o_t^i \) is the agent’s observation, defined as an arbitrary transformation of the environment state \( s_t \). In our case, \( o_t^i \) is different and contains specific information for each agent. The joint action \( \mathbf{a}_t = (a_t^1, a_t^2, \dots, a_t^n) \) determines the next environment state \( s_{t+1} \) based on \( P \) and \( s_t \), and each agent receives a reward \( r_t^i = r^i(s_t, \mathbf{a}_t) \). Lastly, the agents aim to maximize their discounted cumulative return:
\begin{equation}
G_t^i = \sum_{k=0}^{\infty} \gamma^k r_{t+k+1}^i.
\end{equation}

Since agents interact in a shared environment, their actions are interdependent, requiring strategies that consider both individual and collective behaviors. Learning in \gls{marl} thus involves finding policies \( \pi^i \) that optimize rewards while accounting for coordination, competition, or cooperation among agents.

\subsection{Google Research Football and Training Framework} \label{sec:prelims-grf}
\gls{grf} is a physics-based 3D football simulation implemented on top of an existing football game engine~\cite{kurach2020gfootball}. It provides both single and multi-player control options, as well as stochasticity in, e.g., shots, making it appropriate for \gls{rl} and game-AI research. 
Moreover, there are pre-defined scenarios for both benchmarking and curriculum learning that include standard football rules, roles, and match states, while allowing for custom scenarios. In this work, we create our own 11-vs-11 scenarios for the curriculum and subsequent learning phases, as described in Section~\ref{sec:experiments}. For training, we use the distributed training framework for \gls{grf} by Song et al. \cite{song2023boosting}, which is tailored to self-play and multi-agent methods. Furthermore, \gls{grf} contains a rule-based \acrfull{hai}, which actively controls the selected player while applying basic football behavior to non-active ones. \gls{hai} accepts a configurable difficulty factor, where higher values correspond to faster decision-making and reaction times. We use this difficulty factor to adjust our curriculum and employ \gls{hai} as the baseline for evaluations (see Section~\ref{sec:experiments}).

\minisection{Observation Space.} We use customized observation vectors for the actor and critic components (see Section~\ref{sec:methodology-tizero}) following TiZero's design, which is based on \gls{grf}'s floats observation vector, further augmented with more football-specific features, e.g., offside flags \cite{lin2023tizero,kurach2020gfootball}. In total, there are $330$ float observation values for the actor and $220$ for the critic. The elements of the observation space for the actor and critic are listed in more detail in Appendix~\ref{app:observation-elements}.

\minisection{Action Space.} In this work, we use the action set provided by \gls{grf} for the individual agents. These actions include movement in eight directions and various ways to interact with the ball, e.g., shoot or a high pass. Additionally, the agents can sprint (affecting their tiredness), slide tackle, or dribble. In total, we consider 18 discrete actions for each of our agents and exclude \gls{grf}'s ``Built-in AI'' action (which transfers control to the \gls{hai}) from their action set.
\looseness=-1

\minisection{Reward Function.} \label{sec:tizero-reward-function}
\gls{grf} provides two built-in reward functions, and in this work, we use the ``SCORING'' function which returns $ +1 $ for scoring a goal and $ -1 $ for conceding one. Moreover, based on TiZero's design \cite{lin2023tizero}, we add to this built-in reward four additional dense signals extracted from the environment: a ``Hold-Ball Reward'' ($ +0.0001 $ per time-step when a teammate controls the ball), a ``Passing-Ball Reward'' ($+0.05$ for a successful pass), a ``Grouping Penalty'' ($-0.00002$ for each teammate pair getting too close and, consequently, $-0.001$ when the whole team clusters, considering a closeness threshold of $0.05$ distance units), and an ``Out-of-Bounds Penalty'' ($ -0.001$ when an agent moves outside the field boundaries). These constitute our base TiZero reward function, which we extend further in Section~\ref{sec:methodology}. Finally, this reward function is zero-sum across the two teams, i.e., a positive reward for a team gives the equivalent negative for the other.

\section{Methodology} \label{sec:methodology}
In this section, we first describe TiZero as the basis of our method \cite{lin2023tizero}, and afterwards, the architectural modifications and other minor variations from the original approach with the aim of computational efficiency. We then explain our proposed augmentation by utilizing the \acrfull{rnd} exploration term in order to improve sample efficiency.

\subsection{TiZero} \label{sec:methodology-tizero}
TiZero is a \gls{marl} method using curriculum learning and self-play to train team-based football-gameplay policies \cite{lin2023tizero}. The method utilizes an actor-critic paradigm where agents are trained together using a global reward signal and evaluated for actions individually, following the \gls{ctde} concept. Specifically, TiZero uses a shared actor network for individual agents, differentiating between them by taking player-specific information as input. Moreover, TiZero utilizes a policy loss in the form of \gls{jrpo}, which is a modification of \gls{mappo} \cite{yu2022mappo_ippo}. Specifically the \gls{jrpo} loss considers a joint policy over the individual policies:
\begin{equation}
    \pi_{\theta}(\mathbf{a}_t | \mathbf{o}_t) \approx \prod_{i=1}^{n} \pi_{\theta}^i(a_t^i | o_t^i), 
\end{equation}
where $ \pi_{\theta}^i(a_t^i | o_t^i) $ denotes the individual policy for agent $ i $, and $ \mathbf{a}_t $ and $ \mathbf{o}_t $ denote the joint action and observations for all agents. This joint policy is then used in the loss below to create the full \gls{jrpo} actor loss:
\begin{equation} 
\label{eq:total-polic-loss}
\begin{aligned}
L^{\text{JRPO}}(\theta) = \hat{\mathbb{E}}_t \left[ \min \left( \rho_t(\theta) \hat{A}_t, \Pi_{[1 - \epsilon, 1 + \epsilon]}(\rho_t(\theta)) \hat{A}_t \right) \right] \\
+ \quad \beta \hat{\mathbb{E}}_t\left[ H_{\pi_{\theta}} \right],
\end{aligned}
\end{equation}
where $ \rho_t(\theta) $ is the ratio between the new and old joint policies, $ \hat{A}_t $ is the global advantage estimation computed via generalized advantage estimation~\cite{schulman2015gae}, $ \Pi_{[1 - \epsilon, 1 + \epsilon]}(\cdot) $ is a projection operator which clips its argument to the range \( [1 - \epsilon, 1 + \epsilon] \) with $ \epsilon $ as a hyper-parameter, and $ H_{\pi_{\theta}}$ is the entropy of the current joint policy with $ \beta $ as its weighting hyper-parameter. Additionally, the critic uses a \acrfull{mse} loss, optimizing its value estimation towards a global reward signal for a given environment state.

TiZero trains the final joint policy from scratch through a two-stage pipeline of first curriculum learning and then self-play, which is further divided into two sub-stages: \textit{Challenge} and \textit{Generalize}, inspired by OpenAI 5 and AlphaStar, respectively~\cite{openai2019dota, vinyals2019grandmaster_league_sp}. The purpose of the curriculum stage is to develop a basic understanding of simple tactics through a progression of football scenarios with increasing difficulties (e.g., moving towards the goal to shoot and score), whereas the self-play stage improves gameplay on top of this basic knowledge. 

In self-play, TiZero improves its performance through a two-step training approach. In the \textit{Challenge} phase, it primarily trains against the most recently-learned joint policy, improving strategy by further optimizing on the previously-learned behavior. In the \textit{Generalize} phase, it mainly trains against older versions which the current policy struggles to defeat. This step helps TiZero overcome local optima from earlier training stages and addresses the non-transitivity issue in football policies \cite{czarnecki2020real_non_transitivity}.

\subsection{Architectural Modifications} \label{sec:methodology-arch-mods}
To improve computational efficiency and performance, we propose several architectural changes to TiZero. Firstly, we replace the \gls{lstm} element in TiZero's actor and critic with a 4-layer \gls{mlp}. This is because we are concerned with a traditional and industrial game development setting where time and resources are limited, and so, we aim to reduce the computational complexity of the methods under study as much as possible.
\begin{figure*}
    \centering
    \includegraphics[width=0.9\textwidth]{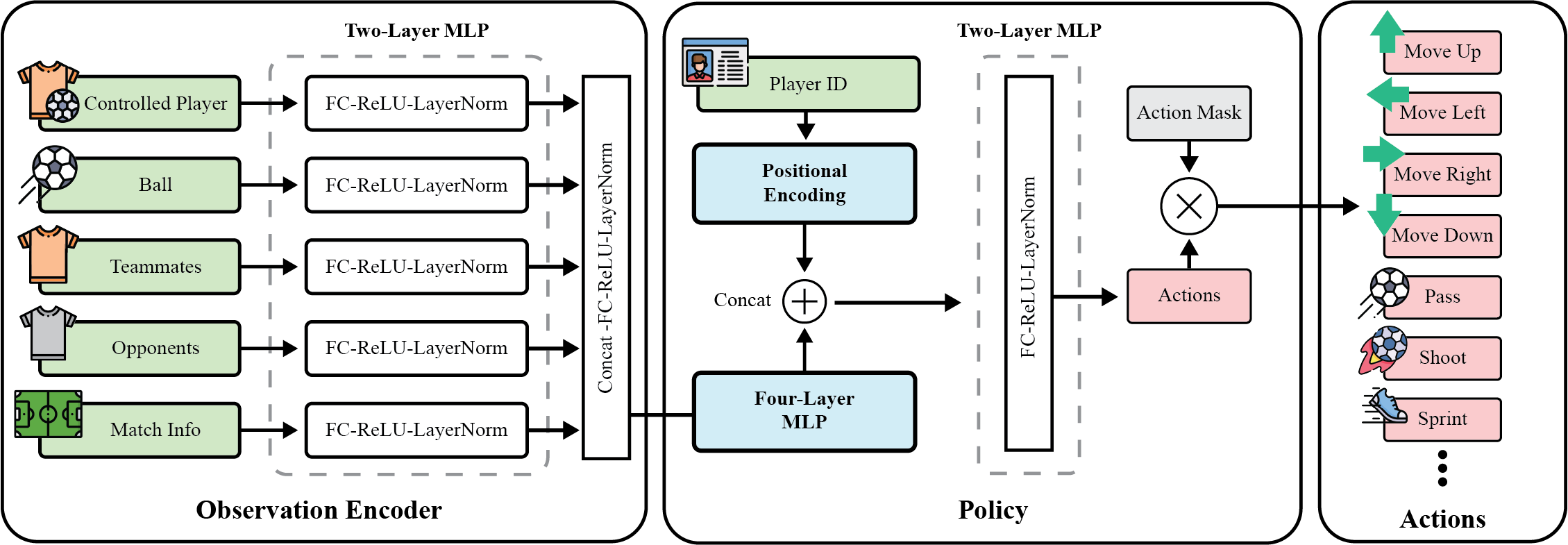}
    \caption{An overview of the architecture of our modified TiZero actor component, with our modifications highlighted in blue. We replace the original \gls{lstm} in TiZero with a 4-layer \gls{mlp} and the original player-ID \gls{mlp} encoder with fixed positional encodings. This network is shared among agents and by passing their player-related information beside the global ones, the network differentiates among individual agents.}
    \Description{The figure shows a left-to-right block diagram with three stages. The Observation Encoder on the left takes five input streams (Controlled Player, Ball, Teammates, Opponents, and Match Info), passes each through a two-layer FC-ReLU-LayerNorm \gls{mlp}, and concatenates them into a single FC-ReLU-LayerNorm embedding. The Policy stage in the middle combines this embedding with a Player ID input that passes through a Positional Encoding block and a Four-Layer \gls{mlp}, both highlighted in blue. The two streams are concatenated and sent through another FC-ReLU LayerNorm. The output is multiplied by an Action Mask to produce the final Actions on the right, such as Move Up, Move Left, Move Right, Move Down, Pass, Shoot, and Sprint.}
    \label{fig:our-tizero-arch}
\end{figure*}

Moreover, for the player ID encoding in the actor architecture, we substitute the original \gls{mlp} with the fixed positional encoding used in Transformer architectures and defined in the work from Vaswani et al.~\cite{vaswani2023attention}. Fixed positional encodings often provide a more stable representation of player IDs, helping the shared actor network distinguish agents without introducing additional learnable parameters. We evaluate this approach in Section~\ref{sec:experiments}, where we ablate this technique and compare to the original MLP-based encoding approach. Our final, modified actor network architecture can be seen in Fig.~\ref{fig:our-tizero-arch}. 

We use similar modifications for the critic. Following \gls{mappo}'s approach, we also normalize the value estimate targets for better stability during the critic's training and preventing value explosion or vanishing gradients \cite{yu2022mappo_ippo}.

\subsection{Random Network Distillation}
As an augmentation of TiZero, we include a state exploration bonus using \gls{rnd} \cite{burda2018rnd}. This technique provides a learned curiosity term that rewards novel and rare states. We extend the single-agent framework of Burda et al.~\cite{burda2018rnd} to \gls{marl} by defining a global, shared exploration bonus for all agents using a global state representation as the input to the \gls{rnd} module. More specifically, we use two neural networks: a fixed and randomly-initialized target network $\psi$, and a predictor network $\hat{\psi}$ trained on data collected by the agent. Our \gls{rnd} exploration bonus is then defined as the \acrshort{mse} between these two networks:
\begin{equation}
    r_{\text{c}}(s_t) = (\hat{\psi}(s_{t+1}) - \psi(s_{t+1}))^2,
\end{equation}
where $s_t$ is the global state visited by the agents (i.e., the same one used by our value network) at timestep $t$. The predictor network trains on the randomly-initialized network's output, and the more frequently a state is visited by agents, the lower the \acrshort{mse} prediction error, and consequently, the exploration bonus will be. 
Using this bonus, we define our \gls{rnd}-based reward function:
\begin{equation}
    R_{\text{RND}}(t) = R_{\text{TiZero}}(s_t, a_t, s_{t+1}) + r_{\text{c}}(s_t).
\end{equation}

We use \glspl{mlp} with \acrshort{relu} activations for both target and predictor networks, with the predictor (four layers) having more layers than the target (two layers). Both networks output four values, with this output dimension treated as a hyper-parameter. We believe combining \gls{rnd} with the entropy term in the actor loss (see Eq.~\ref{eq:total-polic-loss}) provides a balanced form of exploration, since \gls{rnd} encourages exploration by rewarding novel joint states, while the entropy term maintains policy stochasticity, promoting diverse action selection.

\section{Experiments} \label{sec:experiments}

In this section, we present our training setup, evaluation settings, and results for TiZero and our augmented version using \gls{rnd}. We first describe our curriculum learning and self-play setup, followed by the settings of our training experiments and evaluations. We then present results for the training experiments and analyze the sample efficiency of each method. Finally, we evaluate the generalization of the best trained models against a \gls{hai} baseline and provide detailed gameplay statistics and analysis. 

Note that our main goal in this work is to improve the sample efficiency of TiZero as a representative \gls{marl} method. We conduct training experiments on both the original TiZero and TiZero augmented with \gls{rnd} within a fixed data sample budget. Therefore, since we are not concerned with improving the gameplay quality of the final models, we do not run the methods for $40$ days as was done in the original TiZero paper \cite{lin2023tizero}. However, we still carry out gameplay evaluations against a baseline \gls{hai} to test the generalization ability of our resulting models.

\subsection{Experimental Setup} \label{sec:training-exp-setup}
All training experiments use 5 different seeds and are carried out on machines equipped with an Intel Xeon Gold 6146 CPU @ 3.20GHz (12 cores, 24 logical processors), 256GB of DDR4 RAM, and an NVIDIA GeForce GTX 1080 Ti GPU with 11GB of VRAM.

\begin{table*}
\centering
\caption{Sample-efficiency metrics for the curriculum, \textit{Challenge}, and \textit{Generalize} stages of training in \textbf{TiZero} and \textbf{TiZero-RND}. Interquartile mean values with standard deviations over 5 seeds are used, and elapsed time is in hours. N/A indicates that the value for the corresponding row and column is not available due to the method not reaching the respective training stage.}
\setlength{\tabcolsep}{4pt} 
\renewcommand{\arraystretch}{1.2} 
\scalebox{0.75}{ 
\begin{tabular}{l|S[table-format=2.2(1.2)]|S[table-format=3.2(3.2)]|S[table-format=1.2(1.2)]|S[table-format=1.2(1.2)]|S[table-format=4.2(4.2)]|S[table-format=2.2(2.2)]|S[table-format=1.2(1.2)]|S[table-format=4.2(4.2)]|S[table-format=2.2(2.2)]}
        \textbf{Method} & 
        \shortstack{\textbf{Passed} \\ \textbf{Curriculum} \\ \textbf{Scenarios $\uparrow$}} & 
        \shortstack{\textbf{\textit{Curriculum}} \\ \textbf{Rollouts $\downarrow$}} & 
        \shortstack{\textbf{\textit{Curriculum}} \\ \textbf{Elapsed Time $\downarrow$}} & 
        \shortstack{\textbf{Passed} \\ \textbf{\textit{Challenge}} \\ \textbf{Phases $\uparrow$}} & 
        \shortstack{\textbf{\textit{Challenge}} \\ \textbf{Rollouts $\downarrow$}} & 
        \shortstack{\textbf{\textit{Challenge}} \\ \textbf{Elapsed Time $\downarrow$}} & 
        \shortstack{\textbf{Passed} \\ \textbf{\textit{Generalize}} \\ \textbf{Phases $\uparrow$}} & 
        \shortstack{\textbf{\textit{Generalize}} \\ \textbf{Rollouts $\downarrow$}} & 
        \shortstack{\textbf{\textit{Generalize}} \\ \textbf{Elapsed Time $\downarrow$}} \\
        \hline
        \bfseries TiZero \cite{lin2023tizero} & \bfseries 10.00 \pm 0.00 & 275.30 \pm 19.57 & 5.63 \pm 0.38 & \bfseries 1.00 \pm 0.00 & 2940.33 \pm 330.98 & 60.15 \pm 6.19 & {N/A} & {N/A} & {N/A} \\
        \bfseries TiZero-RND (Ours) & \bfseries 10.0 \pm 0.00 & \bfseries 261.87 \pm 50.58 & \bfseries 5.36 \pm 1.06 & 0.5 \pm 0.5 & \bfseries 2297.00 \pm 486.19 & \bfseries 47.02 \pm 9.88 & \bfseries 0.25 \pm 0.43 & \bfseries 978.0 \pm 111.0 & \bfseries 19.78 \pm 2.24 \\
\end{tabular}}
\label{tab:training-exp-results}
\end{table*}

\minisection{Curriculum and Self-Play Design. }
Following TiZero, we employ 10 scenarios with increasing difficulties in our curriculum, where in each scenario, the agents play against the last version of the trained policy in the previous scenario (and \gls{grf}'s heuristic AI in the first scenario). In each scenario, the agents progress to the next one when they reach a win-rate threshold assigned to the curriculum scenario. 

In our curriculum design, we do not use a constant win-rate threshold as done in TiZero. Instead, we set a starting win-rate threshold in the curriculum stage at 55\% (slightly above 50\% for confidence) for scenario~1 and increase it linearly to 75\% by scenario~8. We do so because we found an uneven difficulty gap between curriculum scenarios, specifically from scenario 7 to 8, during the curriculum learning stage of TiZero. We believe the agents spend an unnecessarily long time on simpler scenarios in TiZero's curriculum design, potentially overfitting to basic strategies (e.g., moving into the goal with the ball) that are not particularly effective at later scenarios. Furthermore, we lower the maximum passing win-rate threshold in the curriculum stage from TiZero's 80\% to 75\%, aiming to maintain a similar performance level while reducing training time requirements. Since the original TiZero paper does not specify a win-rate threshold for their curriculum, we assumed it matched the 80\% used in their \textit{Challenge} and \textit{Generalize} stages. 

Each curriculum scenario consists of an 11-vs-11 football game with offside disabled. The game lasts at most \( 500 \) environment steps but ends immediately if a foul is committed, a player goes out of bounds, or a team scores and is deemed the winner of the scenario. In all cases of termination, the environment is reset. Moreover, the players' positioning is adjusted in each scenario following TiZero's approach \cite{lin2023tizero}. However, in our design, we manually set the \gls{grf} player strength factors to \texttt{[0.05, 0.1, 0.15, 0.2, 0.3, 0.4, 0.5, 0.6, 0.75, 0.95]} from scenario 1 to 10. We take this approach since we observe in TiZero that a linear increase in \gls{grf}'s individual player strength factors (affecting their reaction times) does not provide a linear difficulty increase, resulting in uneven difficulty jumps between curriculum scenarios.

After the curriculum stage, the method enters the self-play stage where it alternates between \textit{Challenge} and \textit{Generalize} phases. For the \textit{Challenge} phase, we use a normal football scenario including offside and ending after \( 3000 \) environment steps, no longer ending on one team scoring but continuing until the end of the match. The opponent is chosen from a pool of previously trained policies using the approach of OpenAI 5 \cite{openai2019dota}, and the agents move to the \textit{Generalize} phase only after passing a fixed win-rate threshold. We set this win-rate threshold to 75\%, as opposed to the 80\% threshold believed to be used in TiZero, aiming to reduce the required training time while retaining a reasonable performance level. We use the same game scenario and threshold for the \textit{Generalize} phase, and the opponent is chosen from the pool of previous policies via the approach of AlphaStar \cite{vinyals2019grandmaster_league_sp}. After passing this stage, training goes back to the \textit{Challenge} phase where it continues the alternation between \textit{Generalize} and \textit{Challenge} phases until manually stopped. Here, training can be stopped if convergence is observed in the performance (e.g. average win-rate) or, if existing, a maximum number of training steps is reached.

\minisection{Heuristic Football Game-AI Baseline.}
As for the scripted game-AI comparison baseline, we have chosen the built-in, scripted football game-AI provided for the \gls{grf} environment in the framework presented by \citet{song2023boosting}. This baseline is essentially a rule-based bot proposed by \citet{konings_bazkiebumpercargameplayfootball_2024_engine}, and it accepts a difficulty level which can be between 0 and 1, where a higher difficulty level corresponds to a faster decision-making and reaction time. Additionally, the non-active players are controlled by another rule-based bot which follows standard, basic football behavior patterns, such as moving towards the ball when not possessing it. This built-in AI can be replaced by another mechanism or disabled in \gls{grf}. Here, we utilize it as a baseline for our agents to play against.

\minisection{Experiments and Baseline Evaluations. } 
We run training experiments on the two methods that we study in this work: TiZero without any intrinsic/exploration reward (TiZero) and TiZero augmented with \gls{rnd} (TiZero-RND). Each model is trained for 170M environment steps, and the intrinsic/exploration reward term is added after $700$ rollouts for network warm up \cite{devidze2022exploration_based_reward_shaping}, where a rollout refers to a fixed, limited number of transitions. This corresponds to $ 0.14 \times 10^8 $ environment steps, as each rollout is performed by a batch of 40 workers, with each worker performing 500 environment steps. On average, a single training session runs for about a week.

Furthermore, we evaluate the generalization of the best trained models from TiZero and TiZero augmented with \gls{rnd} by having them play a number of football games against a \gls{hai} baseline provided by \gls{grf} and described in Section~\ref{sec:prelims-grf}. It should be noted that the agents optimize against this baseline \gls{hai} only in the first curriculum scenario and spend most of the training in playing against versions of themselves \cite{lin2023tizero}. Therefore, we are effectively evaluating the generalization of the methods under study. For these evaluations, the football scenario used is the same as the one for \textit{Challenge} and \textit{Generalize} phases, i.e., a normal football match scenario. We report gameplay performance metrics for each method averaged over \( 50 \) evaluation matches, repeating the experiment with 5 different seeds. We use the \gls{hai} baseline with the medium strength level.\looseness=-1

Lastly, we perform an ablation study of the effect of the fixed positional encoding for player ID embedding compared to TiZero's approach. In the original design, one-hot encoding is used for a given player ID that is then input to a fully-connected layer, outputting a number of latent values equal to the length of the original one-hot encoding \cite{lin2023tizero}. As we replace this part of the architecture with a fixed positional encoding, we perform training experiments on TiZero-RND using the original MLP-based encoding for the player IDs and compare the results to the ones for TiZero-RND with positional encoding used. We thereby analyze the impact of using the fixed positional encoding approach on training sample efficiency and stability compared to the original MLP encoding.  

Appendix~\ref{app:exp-configs-hyperparams} presents a comprehensive list of the hyperparameters used in our experiments. In addition, Appendix~\ref{app:ssir-method} reports supplementary experiments that use an alternative exploration strategy based on a different form of intrinsic reward. Specifically, we replace the RND module with the \Gls{ssir}~\cite{devidze2022exploration_based_reward_shaping}, and we refer the reader to the appendix for a detailed description of this methodology and its corresponding results.

\subsection{Results}
We first discuss the sample efficiency of the methods during training and then evaluate the gameplay generalization of the best models of each method against a scripted heuristic AI as a baseline (see Section~\ref{sec:training-exp-setup}). Furthermore, we analyze the effect of using the fixed positional encoding for player ID embedding compared to TiZero's original approach via an ablation study.

\minisection{Training Sample Efficiency. } We train each of the methods as described in Section~\ref{sec:training-exp-setup}. We summarize the results with respect to sample-efficiency metrics in Table~\ref{tab:training-exp-results}, namely the number of training phases passed, the elapsed time and the number of rollouts required, on average, to pass a particular phase. We have derived these metrics for the curriculum, \textit{Challenge}, and \textit{Generalize} stages. It should be noted that progress in the training process is mainly implied by the number of training phases passed, as the goal of the agents is to achieve given win-rate thresholds at each phase to proceed to the next.

From Table~\ref{tab:training-exp-results}, we see that both TiZero and TiZero-RND pass all of the curriculum scenarios. Both exhibit a similar amount of elapsed time and number of rollouts in the curriculum phase as well, while TiZero-RND appears to require slightly fewer rollouts here. More specifically, TiZero-RND achieves a $4.8\%$ improvement considering the average elapsed time required for each method to pass a curriculum scenario. In the \textit{Challenge} phase, TiZero passes more stages on average; however, TiZero-RND demands a lower number of rollouts and elapsed time to go through the \textit{Challenge} stages. Similarly to the curriculum phases, in these stages TiZero-RND achieves an improvement rate of $21.8\%$ when comparing each method's average elapsed time per \textit{Challenge} stage.

Furthermore, we observe that standard TiZero does not complete any stages in the \textit{Generalize} phase, whereas TiZero-RND does. Additionally, TiZero-RND attains its number of \textit{Generalize} stages with a number of rollouts and elapsed time which is lower than the ones needed in the \textit{Challenge} phase for both standard TiZero and TiZero-RND. Considering that the \textit{Generalize} phase is a difficult phase where the models are faced with versions of themselves against whom they had previously demonstrated worse performance, we believe TiZero-RND has better utilized its training data within our fixed training budget to allow for a better generalization. We mainly attribute this better utilization of training data to its more efficient exploration of the game states via the added \gls{rnd} term during training.

\begin{figure}
  \centering
  \subfigure[Win-Rate Comparison During Training]{%
    \includegraphics[width=\columnwidth]{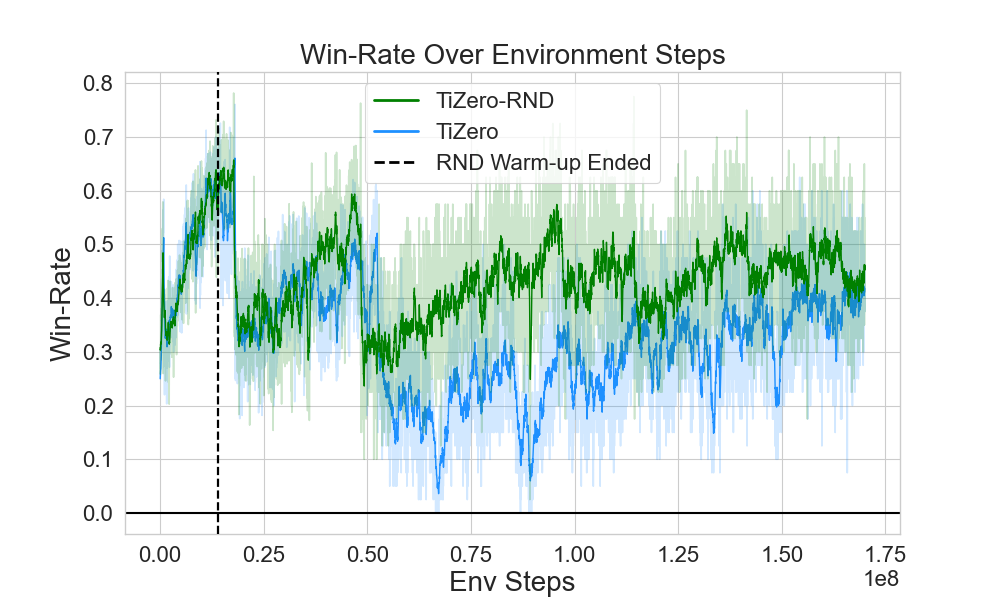}%
    \Description{A line plot of win-rate against environment steps, ranging from 0 to $1.75 \times 10^8$ on the x-axis and 0 to 0.8 on the y-axis. The green curve shows TiZero-RND and the blue curve shows standard TiZero, with a vertical dashed line marking the end of the RND warm-up phase near $0.15 \times 10^8$ steps. Both methods climb to a win-rate of about 0.6 during warm-up and then drop sharply. After the warm-up ends, TiZero-RND recovers and oscillates between 0.4 and 0.5 for the remainder of training, while standard TiZero stays lower and noisier, oscillating between 0.2 and 0.4. Shaded regions around each curve show the raw values.}
    \label{fig:best-rnd-vs-best-tizero-win}%
  }
  \subfigure[Reward Comparison During Training]{%
    \includegraphics[width=\columnwidth]{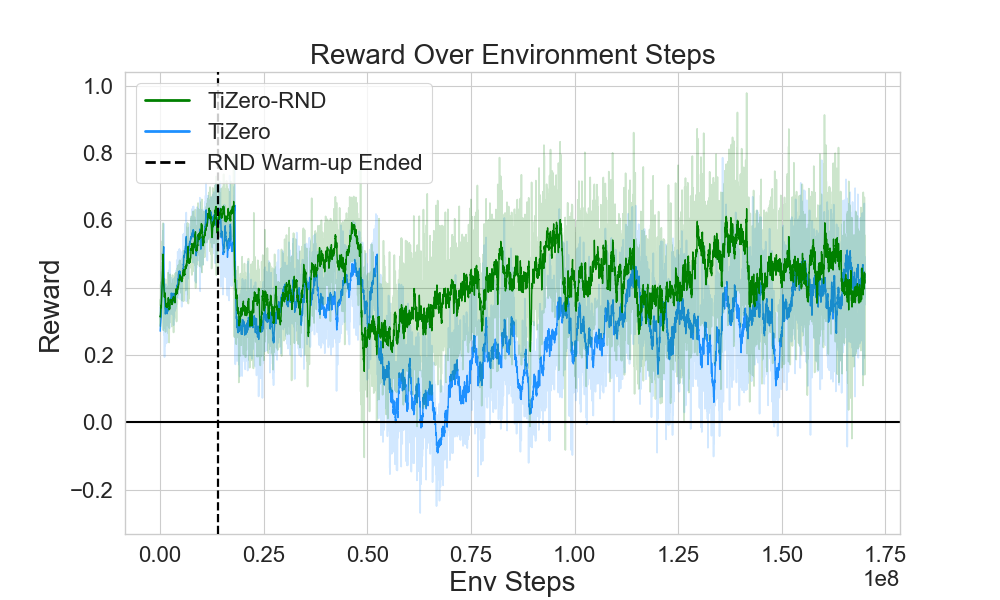}
    \Description{A line plot of reward against environment steps, ranging from 0 to $1.75 \times 10^8$ on the x-axis and about -0.2 to 1.0 on the y-axis. The green curve shows TiZero-RND and the blue curve shows standard TiZero, with a vertical dashed line marking the end of the RND warm-up phase near $0.15 \times 10^8$ steps. Both methods reach a reward of about 0.6 during warm-up and then drop, with standard TiZero briefly going negative. After the warm-up ends, TiZero-RND recovers and oscillates between 0.4 and 0.5, while standard TiZero remains lower and noisier, oscillating between 0.2 and 0.4. Shaded regions around each curve show the raw values.}%
    \label{fig:best-rnd-vs-best-tizero-reward}%
  }
  \caption{Comparison of win-rate and reward achieved by the best-performing TiZero-RND (green) and the best-performing standard TiZero (blue) experiments during training. The solid points represent the exponential moving average, and the shaded points are the raw values.}
  \label{fig:rnd-vs-tizero-training-plots}
\end{figure}
\begin{table*}
\centering
\caption{Gameplay metrics against a baseline \acrfull{hai}, which is less-seen during training, for \textbf{TiZero} and \textbf{TiZero-RND}. Interquartile means and standard deviations are computed over 50 games per best model, aggregated over 5 seeds, and the baseline is used on ``medium'' difficulty. Good and bad passes and shots are calculated as percentages of the total metric value.}
\setlength{\tabcolsep}{4pt} 
\renewcommand{\arraystretch}{1.2} 
\scalebox{0.93}{ 
\begin{tabular}{l
                |S[table-format=2.2(2.2)]
                |S[table-format=2.2]
                |S[table-format=2.2]
                |S[table-format=1.2(1.2)]
                |S[table-format=2.2]
                |S[table-format=3.2]
                |S[table-format=3.1(3.1)]
                |S[table-format=2.1(1.1)]
                |S[table-format=2.1(1.1)]}
    \textbf{Method} & 
    \shortstack{\textbf{Total} \\ \textbf{Passes}} & 
    \shortstack{\textbf{Good} \\ \textbf{Pass \%} $\uparrow$} & 
    \shortstack{\textbf{Bad} \\ \textbf{Pass \%} $\downarrow$} &
    \shortstack{\textbf{Total} \\ \textbf{Shots}} &
    \shortstack{\textbf{Good} \\ \textbf{Shot \%} $\uparrow$} & 
    \shortstack{\textbf{Bad} \\ \textbf{Shot \%} $\downarrow$} & 
    \shortstack{\textbf{Total} \\ \textbf{Possession}} &
    \shortstack{\textbf{Interception} $\uparrow$} &
    \shortstack{\textbf{Get} \\ \textbf{Intercepted} $\downarrow$} \\
    \hline
    TiZero \cite{lin2023tizero} & \bfseries 31.32 \pm 21.82 & \bfseries 65.85 & \bfseries 33.30 & \bfseries 0.66 \pm 0.68 & 5.42 & 92.77 & \bfseries 646.1 \pm 218.3 & 13.2 \pm 1.2 & \bfseries 12.2 \pm 2.7  \\
    \bfseries TiZero-RND (Ours) & 28.66 \pm 6.35 & 60.61 & 40.09 & 0.11 \pm 0.12 & \bfseries 18.18 & \bfseries 72.73 & 619.7 \pm 37.2 & \bfseries 13.6 \pm 0.2 & 13.0 \pm 0.8  \\
\end{tabular}}
\label{tab:detailed-gameplay-eval-metrics}
\end{table*}
 
In conclusion, TiZero-RND improves sample efficiency per training stage based on the number of rollouts and elapsed time it requires to pass through a single stage, specifically by $13.3\%$ considering both curriculum and Challenge phases, and it also improves generalization over long training experiments compared to the original TiZero, as it manages to succeed and adapt better against opponents that previously yielded poor performance on average.

Lastly, Fig.~\ref{fig:rnd-vs-tizero-training-plots} shows comparison plots of the win-rate and reward values achieved by the best-performing TiZero-RND and the best-performing TiZero experiments (among their respective $5$ seeds). In the figure, both methods exhibit a similar trend throughout training with peaks followed by sudden drops, indicating a curriculum or stage change. This is while the trajectory within each given training phase (i.e., after drops) remains upward. Nevertheless, TiZero-RND appears to achieve a higher win-rate and reward values throughout most of the training session. Additionally, while both methods experience drops especially when the training stage changes (i.e., often due to increased difficulty), TiZero-RND's drops seem to be less pronounced compared to the ones for standard TiZero. For instance, between the 50 and 100 million environment steps marks, the win-rate for standard TiZero drops near zero at times, while TiZero-RND's value remains stable between $0.3$ and $0.6$. We believe these results indicate a steadier learning performance for TiZero-RND, possibly due to its enhanced exploration.
\begin{table*}[t]
\centering
\caption{Ablating the fixed positional encoding for player ID embedding and using the original MLP approach. Interquartile mean values with standard deviations over 5 seeds are used, and elapsed time is in hours.}
\setlength{\tabcolsep}{4pt} 
\renewcommand{\arraystretch}{1.2} 
\scalebox{0.72}{ 
\begin{tabular}{l|S[table-format=2.2(1.2)]|S[table-format=3.2(3.2)]|S[table-format=1.2(1.2)]|S[table-format=1.2(1.2)]|S[table-format=4.2(4.2)]|S[table-format=2.2(2.2)]|S[table-format=1.2(1.2)]|S[table-format=4.2(4.2)]|S[table-format=2.2(2.2)]}
        \textbf{Method} & 
        \shortstack{\textbf{Passed} \\ \textbf{Curriculum} \\ \textbf{Scenarios $\uparrow$}} & 
        \shortstack{\textbf{\textit{Curriculum}} \\ \textbf{Rollouts $\downarrow$}} & 
        \shortstack{\textbf{\textit{Curriculum}} \\ \textbf{Elapsed Time $\downarrow$}} & 
        \shortstack{\textbf{Passed} \\ \textbf{\textit{Challenge}} \\ \textbf{Phases $\uparrow$}} & 
        \shortstack{\textbf{\textit{Challenge}} \\ \textbf{Rollouts $\downarrow$}} & 
        \shortstack{\textbf{\textit{Challenge}} \\ \textbf{Elapsed Time $\downarrow$}} & 
        \shortstack{\textbf{Passed} \\ \textbf{\textit{Generalize}} \\ \textbf{Phases $\uparrow$}} & 
        \shortstack{\textbf{\textit{Generalize}} \\ \textbf{Rollouts $\downarrow$}} & 
        \shortstack{\textbf{\textit{Generalize}} \\ \textbf{Elapsed Time $\downarrow$}} \\
        \hline
        TiZero-RND (w/o PE) & \bfseries 10.00 \pm 0.00 & 352.90 \pm 76.15 & 7.29 \pm 1.59 & 0.40 \pm 0.49 & \bfseries 1739.50 \pm 69.75 & \bfseries 36.21 \pm 1.26 & \bfseries 0.40 \pm 0.49 & 2568.00 \pm 566.00 & 53.19 \pm 12.00 \\
        \bfseries TiZero-RND (w/ PE, Ours) & \bfseries 10.0 \pm 0.00 & \bfseries 261.87 \pm 50.58 & \bfseries 5.36 \pm 1.06 & \bfseries 0.5 \pm 0.5 & 2297.00 \pm 486.19 & 47.02 \pm 9.88 & 0.25 \pm 0.43 & \bfseries 978.0 \pm 111.0 & \bfseries 19.78 \pm 2.24 \\
\end{tabular}}
\label{tab:ablation-player-id-embedding}
\end{table*}

\minisection{Generalization vs. Baseline Heuristic AI. } To further evaluate the gameplay performance and generalization to less encountered opponents, we run $50$ games against the baseline \gls{hai} (see Section~\ref{sec:training-exp-setup}) for the top-performing models of the original TiZero and TiZero-RND. We report football-specific gameplay metrics in Table~\ref{tab:detailed-gameplay-eval-metrics} to assess the overall behavioral quality of the models, including the percentage of good and bad passes and shots, total possession, and interception counts. Specifically, according to \gls{grf}, a good pass is one that successfully reaches a teammate, while a bad pass fails to do so. Similarly, a good shot results in a goal, whereas a bad shot is one towards the goal which does not score.

In passing, both methods attempt a comparable number of total passes, while TiZero achieves a higher percentage of good passes. However, a significant divergence is observed in the shooting behaviors; while TiZero-RND attempts much fewer total shots on average, it demonstrates a higher accuracy and precision as it attains a higher percentage of good shots and a lower rate of bad shots. Additionally, in possession, both methods showcase a comparable performance, with standard TiZero achieving slightly more possession on average. Defensively, TiZero-RND proves more effective at intercepting the ball, whereas standard TiZero is better at retaining possession without getting intercepted.

Notably, TiZero-RND showcases significantly lower variance across most metrics; for instance, its standard deviation for the number of shots and possession is roughly five to six times lower than that of standard TiZero. Therefore, we observe that TiZero-RND showcases a generalized gameplay similar to standard TiZero while reaching a higher accuracy in shooting. Here, higher shooting accuracy indicates more chances to score a goal. It further appears to have converged to a more certain and stable behavior as demonstrated by the lower standard deviation across reported metrics.

\minisection{Effect of Fixed Positional Encoding on Player ID Embedding. } Here, we ablate the original one-hot MLP encoding for the player ID against our approach of using a fixed positional encoding instead. We therefore compare TiZero augmented with \gls{rnd}, which showed better sample efficiency in our experiments above, in two versions: one using the original MLP encoding for player IDs (i.e., ablated), and another using fixed positional encoding (i.e., as described in Section~\ref{sec:methodology-arch-mods}). To compare, we run training experiments using these two versions in a similar fashion to our previous experiments for TiZero and TiZero-RND in Table~\ref{tab:training-exp-results}.

The sample-efficiency metrics for our training experiments using the ablated and unmodified versions of TiZero-RND are summarized in Table~\ref{tab:ablation-player-id-embedding}. Examining the table, it can be seen that both versions pass all of the curriculum scenarios in their experiments. However, TiZero-RND with positional encoding appears to require slightly less time and number of rollouts to pass the curriculum part of training compared to the ablated version. Regarding the Challenge phases, although the positional-encoding variant passes slightly more phases, the ablated TiZero-RND requires fewer rollouts and time to attain the Challenge phases. On the other hand, this relation is reversed for the Generalize phase; the ablated version goes through more Generalize phases, while TiZero-RND with positional encoding demands substantially fewer rollouts and shorter elapsed time to complete its phases.

Overall, aggregating the average time per phase across all stages in Table~\ref{tab:ablation-player-id-embedding}, we can see that the ablated TiZero-RND requires on average $\sim 108$ hours in total to reach a fixed number of environment steps (170M as described in Section~\ref{sec:training-exp-setup}), whereas the positional-encoding version takes $\sim 82$ hours. Additionally, considering the number of phases passed, neither method clearly outperforms the other, as they both demonstrate comparable progress throughout the training stages. Furthermore, by utilizing a fixed positional encoding, we reduce the total number of learnable parameters in the non-ablated version. Consequently, based on this architectural efficiency, together with the significantly lower wall-clock time to achieve similar results, we prefer TiZero-RND with positional encoding in our ablation study.

\section{Limitations}
\label{sec:limitations}

In this section, we discuss some limitations of our work.
We do not do an exact one-to-one comparison with TiZero as presented by Lin et al. \cite{lin2023tizero}. While the method is outlined in significant detail in the original paper, an implementation for the method and its training scheme is not provided at the time of writing, and so, we had to make assumptions on many smaller details in our re-implementation. Additionally, we had to modify the method to bring its computational costs within our time and resource budget. 

Furthermore, we do not train the model for 40 days as is done for the original TiZero, and so, we can not guarantee that the results presented here would necessarily extend to such a prolonged training period. However, such lengthy training sessions are anyway not feasible in conventional game-AI development.
We also do not carry out an extensive hyperparameter analysis; instead, we chose hyperparameters based on preliminary experiments. We consider this sufficient for the purposes of this paper.

Lastly, our approach can be applied to other similar \gls{marl} frameworks, but the anticipated sample-efficiency improvements in domains other than football-like environments are not guaranteed and require further validation and experiments. We view this as a promising direction for future work.

\section{Conclusion} \label{sec:conclusion}

In this work, we have targeted the need for better sample efficiency in multi-agent reinforcement learning methods within a simulated football context. We let TiZero represent a state-of-the-art example of such, which requires excessively long training times to achieve good gameplay performance. To address this, we first modified the architecture of TiZero for computational efficiency. Then, we proposed an augmented version of it by adding a multi-agent random network distillation exploration term to the original reward function. This was motivated by our initial hypothesis that poor sample efficiency can be remedied by better exploration mechanisms. We then carried out training experiments to compare sample efficiency. We saw that TiZero with random network distillation improves sample efficiency by an average of $13.3\%$ in a multi-stage training setup and enhances adaptability to challenging scenarios compared to original TiZero. Lastly, we carried out gameplay evaluations of the learned policies against a heuristic AI baseline, less-seen in training, to test generalization. Our evaluations showed that the random network distillation version achieves a higher accuracy in shooting compared to standard TiZero, as well as higher stability and certainty in its learned behavior.

\bibliographystyle{ACM-Reference-Format}
\bibliography{references}

\appendix
\begin{table*}[ht!]
\centering
\caption{Sample-efficiency metrics for TiZero-SSIR under the curriculum, \textit{Challenge}, and \textit{Generalize} stages of training, besides the same resulting metrics for TiZero and TiZero-RND for comparison. Interquartile mean values with standard deviations over 5 seeds are used, and elapsed time is in hours. N/A indicates that the value for the corresponding row and column is not available due to the method not reaching the respective training stage.}
\setlength{\tabcolsep}{4pt} 
\renewcommand{\arraystretch}{1.2} 
\scalebox{0.75}{ 
\begin{tabular}{l|S[table-format=2.2(1.2)]|S[table-format=3.2(3.2)]|S[table-format=1.2(1.2)]|S[table-format=1.2(1.2)]|S[table-format=4.2(4.2)]|S[table-format=2.2(2.2)]|S[table-format=1.2(1.2)]|S[table-format=4.2(4.2)]|S[table-format=2.2(2.2)]}
        \textbf{Method} & 
        \shortstack{\textbf{Passed} \\ \textbf{Curriculum} \\ \textbf{Scenarios $\uparrow$}} & 
        \shortstack{\textbf{\textit{Curriculum}} \\ \textbf{Rollouts $\downarrow$}} & 
        \shortstack{\textbf{\textit{Curriculum}} \\ \textbf{Elapsed Time $\downarrow$}} & 
        \shortstack{\textbf{Passed} \\ \textbf{\textit{Challenge}} \\ \textbf{Phases $\uparrow$}} & 
        \shortstack{\textbf{\textit{Challenge}} \\ \textbf{Rollouts $\downarrow$}} & 
        \shortstack{\textbf{\textit{Challenge}} \\ \textbf{Elapsed Time $\downarrow$}} & 
        \shortstack{\textbf{Passed} \\ \textbf{\textit{Generalize}} \\ \textbf{Phases $\uparrow$}} & 
        \shortstack{\textbf{\textit{Generalize}} \\ \textbf{Rollouts $\downarrow$}} & 
        \shortstack{\textbf{\textit{Generalize}} \\ \textbf{Elapsed Time $\downarrow$}} \\
        \hline
        \bfseries TiZero-SSIR (Ours) & 8.0 \pm 0.00 & \bfseries 91.30 \pm 3.91 & \bfseries 1.80 \pm 0.05 & {N/A} & {N/A} & {N/A} & {N/A} & {N/A} & {N/A} \\
        \bfseries TiZero \cite{lin2023tizero} & \bfseries 10.00 \pm 0.00 & 275.30 \pm 19.57 & 5.63 \pm 0.38 & \bfseries 1.00 \pm 0.00 & 2940.33 \pm 330.98 & 60.15 \pm 6.19 & {N/A} & {N/A} & {N/A} \\
        \bfseries TiZero-RND (Ours) & \bfseries 10.0 \pm 0.00 & 261.87 \pm 50.58 & 5.36 \pm 1.06 & 0.5 \pm 0.5 & \bfseries 2297.00 \pm 486.19 & \bfseries 47.02 \pm 9.88 & \bfseries 0.25 \pm 0.43 & \bfseries 978.0 \pm 111.0 & \bfseries 19.78 \pm 2.24 \\
\end{tabular}}
\label{tab:training-exp-results-ssir}
\end{table*}

\begin{table*}
\centering
\caption{Gameplay metrics against a baseline \acrfull{hai}, which is less-seen during training, for \textbf{TiZero-SSIR}, \textbf{TiZero}, and \textbf{TiZero-RND}. Interquartile means and standard deviations are computed over 50 games per best model, aggregated over 5 seeds, and the baseline is used on ``medium'' difficulty. Good and bad passes and shots are calculated as percentages of the total metric value.}
\setlength{\tabcolsep}{4pt} 
\renewcommand{\arraystretch}{1.2} 
\scalebox{0.9}{ 
\begin{tabular}{l
                |S[table-format=2.2(2.2)]
                |S[table-format=2.2, table-space-text-post=\,\%]
                |S[table-format=2.2, table-space-text-post=\,\%]
                |S[table-format=1.2(1.2)]
                |S[table-format=2.2, table-space-text-post=\,\%]
                |S[table-format=2.2, table-space-text-post=\,\%]
                |S[table-format=4.1(3.1)]
                |S[table-format=2.1(1.1)]
                |S[table-format=2.1(1.1)]}
    \textbf{Method} & 
    \shortstack{\textbf{Total} \\ \textbf{Passes}} & 
    \shortstack{\textbf{Good} \\ \textbf{Pass \% $\uparrow$}} & 
    \shortstack{\textbf{Bad} \\ \textbf{Pass \% $\downarrow$}} &
    \shortstack{\textbf{Total} \\ \textbf{Shots}} &
    \shortstack{\textbf{Good} \\ \textbf{Shot \% $\uparrow$}} & 
    \shortstack{\textbf{Bad} \\ \textbf{Shot \% $\downarrow$}} & 
    \shortstack{\textbf{Total} \\ \textbf{Possession}} &
    \shortstack{\textbf{Interception $\uparrow$}} &
    \shortstack{\textbf{Get} \\ \textbf{Intercepted $\downarrow$}} \\
    \hline
    TiZero-SSIR (Ours) & 5.48 \pm 4.47 & 38.50\,\% & 61.50\,\% & 0.20 \pm 0.34 & 0.00\,\% & 100.00\,\% & \bfseries 1008.9 \pm 248.4 & 13.7 \pm 0.6 & \bfseries 7.7 \pm 2.1  \\
    TiZero \cite{lin2023tizero} & \bfseries 31.32 \pm 21.82 & \bfseries 65.84\,\% & \bfseries 34.16\,\% & \bfseries 0.66 \pm 0.68 & 5.45\,\% & 94.55\,\% & 646.1 \pm 218.3 & 13.2 \pm 1.2 & 12.2 \pm 2.7  \\
    TiZero-RND (Ours) & 28.08 \pm 5.69 & 59.86\,\% & 40.14\,\% & 0.11 \pm 0.12 & \bfseries 22.72\,\% & \bfseries 77.28\,\% & 614.2 \pm 42.2 & \bfseries 13.9 \pm 0.2 & 13.3 \pm 1.2  \\
\end{tabular}}
\label{tab:ssir-baseline-evaluation}
\end{table*}
\newpage

\section{Additional Methodology: A Self-Supervised Intrinsic Reward as an Exploration Mechanism} \label{app:ssir-method}

Here, we present an additional methodology on a different approach we also investigate, with the same goal of improving sample efficiency through a varied and hopefully better exploration mechanism. More specifically, instead of using random network distillation and its added term, we look into self-supervised reward shaping and its application as an exploration mechanism here. As a more recent and dynamic concept, self-supervised reward shaping utilizes the agent's interactions with the environment and the original reward as self-supervisory signals to learn an additional, denser reward. A notable work is the exploration-guided reward shaping framework, which integrates learned intrinsic rewards with exploration bonuses to improve learning \cite{devidze2022exploration_based_reward_shaping}. Additionally, the authors provide a derivation and formulation for their learned intrinsic reward, from which we take inspiration here for a \Gls{ssir} in our multi-agent framework.

\subsection{Methodology of SSIR}
Aiming to improve the exploration signal, we augment TiZero's original reward (see Section~\ref{sec:tizero-reward-function}) with an \gls{ssir} term, inspired by the work of Devidze et al.~\cite{devidze2022exploration_based_reward_shaping}.
For our implementation, we use a 2-layer \gls{mlp} with \acrshort{relu} activations for the \gls{ssir} network. The input to this network has the same shape as for our actor (see Fig.~\ref{fig:our-tizero-arch}), and the output has the same shape as the number of actions, but with tanh activations. We utilize the network output corresponding to each agent's taken action as the agent's \gls{ssir} term for that transition \cite{devidze2022exploration_based_reward_shaping}.

While Devidze et al. \cite{devidze2022exploration_based_reward_shaping} present a single-agent formulation of this intrinsic reward, we first calculate the values for each agent as described above, and then, we average the intrinsic reward of the individual agents and set that as our global \gls{ssir}, adhering to the \gls{ctde} paradigm (see Section~\ref{sec:related-work-marl-football}). Our resulting \gls{ssir}-based reward function is:
\begin{equation}
    R_{\text{SSIR}}(t) = R_{\text{TiZero}}(s_t, a_t, s_{t+1}) + \alpha \frac{1}{N} \sum_{i=1}^{N}{r_\phi(o^i_t, a^i_t)},
\end{equation}
where \( R_{\text{TiZero}}(s_t, a_t, s_{t+1}) \) is the original TiZero reward function \cite{lin2023tizero}, \( \alpha \) is a hyper-parameter, \( N \) is the number of agents, and \( r_\phi(o^i_t, a^i_t) \) is the \gls{ssir} term computed for each agent by a neural network with parameters \( \phi \). To optimize \( r_\phi \), we use the empirical update rule for \( \phi \) from Devidze et al. \cite{devidze2022exploration_based_reward_shaping} and apply gradient descent by considering the simplified update term as a loss function.

\subsection{SSIR in Training and Evaluation Against Baseline}
In a similar fashion to Section~\ref{sec:experiments}, we run training experiments using TiZero with the provided \gls{ssir} formulation, and we analyze the results besides those provided for TiZero and TiZero-RND to study the effects of the self-supervised intrinsic reward on training sample efficiency. We also evaluate the best policies trained by TiZero-SSIR against our \gls{hai} comparison baseline using the same setup and process used for evaluating the policies trained by TiZero and TiZero-RND in Section~\ref{sec:experiments}.

The training experiment setup is the same as the one provided in Section~\ref{sec:training-exp-setup}, and we aggregate the results for TiZero-SSIR (i.e., TiZero with the self-supervised intrinsic reward) from 5 seeds. The training results for this new method, besides those for TiZero and TiZero-RND, are provided in Table~\ref{tab:training-exp-results-ssir}.
Looking at the training results, we see that TiZero-SSIR passes slightly fewer curriculum scenarios than TiZero and TiZero-RND on average. Additionally, it requires fewer rollouts to complete each curriculum scenario it does pass compared to the other two methods. However, we also observe that this method does not manage to pass any Challenge or Generalize stages, meaning it gets stuck in the curriculum phase. It potentially overfits to simple strategies such as going into the goal with the ball instead of passing and shooting which is more effective in later scenarios. Therefore, considering the results of TiZero-SSIR, we can see that while certain forms of reward shaping, such as \gls{rnd}, improve sample efficiency, others may not bring such enhancements, although they may aim to provide a form of additional exploration.

Furthermore, the gameplay evaluation results for TiZero-SSIR against the baseline, averaged over 50 games and repeated for 5 seeds, are shown in Table~\ref{tab:ssir-baseline-evaluation}. The results for the evaluation of TiZero and TiZero-RND are also brought alongside the new metrics for ease of comparison. 

Considering the gameplay evaluation metrics in Table~\ref{tab:ssir-baseline-evaluation}, the models learned by TiZero-SSIR show a more possessive strategy with the highest total possession among all and getting intercepted almost half the time compared to the other methods. Additionally, these models carry out the fewest number of passes, and the majority of these passes are inaccurate. Regarding shots, the models produced by TiZero-SSIR do not manage to execute any good shots, and thus, do not score any goals. These together indicate that the TiZero-SSIR approach seems to often reach an extreme policy with regards to possession, which is aligned with the training experiment results provided above; the method does not pass the more difficult curriculum scenarios that may require a strategy generalizing better to more complex scenarios. 

\section{TiZero's Actor and Critic Observation Vectors} \label{app:observation-elements}

Here, we detail the elements of the observation vectors used by the actor and critic in our implementation of TiZero \cite{lin2023tizero}:

\begin{itemize}
    \item Controlled player's ID (\textit{only actor}), represented by 1 value;
    \item Controlled player's information (\textit{only actor}). This part includes sticky actions, current player position, current direction, tired factor, yellow card, red card, offside, relative ball position, distance to ball, relative teammate positions, distance to teammates, relative opponent positions, and distance to opponent, for a total of 87 values;
    \item Ball information (\textit{only actor}). This part includes the ball position, ball direction, ball-owning team, ball rotation, ball-owning player, current player information, ball-owning player position, ball-owning player direction, relative position of the ball owner, distance to the ball owner, and ball owner's information, for a total of 57 values.
    \item Teammates' information (\textit{shared for actor and critic}). This part includes teammate positions, teammate directions, teammate tired factors, teammate yellow cards, teammate red cards, and teammate offsides, for a total of 88 values.
    \item Opponents' information (\textit{shared for actor and critic}). This part includes opponent positions, opponent directions, opponent tired factors, opponent yellow cards, opponent red cards, and opponent offsides, for a total of 88 values.
    \item Current match information (\textit{shared for actor and critic}). This part includes the game mode, goal differences, and remaining environment steps, for a total of 9 values.
    \item Ball information (\textit{only critic}). This part includes the ball position, ball direction, ball rotation, and the ball-owning team, for a total of 12 values.
    \item Ball-owning player as a one-hot encoded player index (\textit{only critic}), for a total of 23 values.
\end{itemize}

\section{Experimental Configurations and Hyperparameters} \label{app:exp-configs-hyperparams}
Here, we present the configurations and hyperparameters used for our training experiments. 
Tables~\ref{tab:tizero_rnd_training_config}, \ref{tab:tizero_training_config}, and \ref{tab:tizero_ssir_training_config} provide the training configurations for TiZero-RND, TiZero, and TiZero-SSIR, respectively.

\begin{table}[H]
\centering
\scalebox{1.1}{
\begin{tabular}{|l|l|}
\hline
\textbf{Attribute} & \textbf{Value} \\ \hline
\multicolumn{2}{|l|}{\textbf{Observation Encoder (Actor \& Critic)}} \\ \hline
Input MLP Layers & 2 \\
Input MLP Hidden Units & 64 \\
\textit{Match Info} MLP Hidden Units & 9 \\
Final FC Hidden Units & 256 \\ \hline
\multicolumn{2}{|l|}{\textbf{Actor Policy Network}} \\ \hline
Main MLP Layers & 4 \\
Main MLP Hidden Units & 256 \\
Positional Encoding Dimension & 256 \\
Pre-Action MLP Layers & 2 \\
Pre-Action MLP Hidden Units & 256 \\
Activation Function & ReLU \\ \hline
\multicolumn{2}{|l|}{\textbf{Critic Network}} \\ \hline
Final Encoder Layers & 2 \\
Final Encoder Hidden Units & 256 \\
Activation Function & ReLU \\ \hline
\multicolumn{2}{|l|}{\textbf{Random Network Distillation (RND)}} \\ \hline
Predictor Network Layers & 4 \\
Predictor Network Hidden Units & 128 \\
Target Network Layers & 2 \\
Target Network Hidden Units & 64 \\
Output Embedding Size & 4 \\
Learning Rate & 0.0005 \\
Update Frequency (Rollouts) & 1 \\
Term Coefficient & 0.0001 \\
Start After (Rollouts) & 700 \\ \hline
\multicolumn{2}{|l|}{\textbf{Training Parameters}} \\ \hline
Discount Factor ($\gamma$) & 0.999 \\
GAE Lambda ($\lambda$) & 0.995 \\
PPO Epochs & 2 \\
Num. Mini-batches & 1 \\
Clipping Parameter ($\epsilon$) & 0.2 \\
Value Loss Coefficient & 1.0 \\
Entropy Coefficient & 0.01 \\
Use Orthogonal Initialization & True \\
Initialization Gain & 0.01 \\
Use Feature Normalization & True \\ \hline
\end{tabular}
}
\caption{Configurations and hyperparameters for TiZero-RND's training experiments.}
\label{tab:tizero_rnd_training_config}
\end{table}
\begin{table}[H]
\centering
\scalebox{1.0}{
\begin{tabular}{|l|l|}
\hline
\multicolumn{2}{|l|}{\textbf{Observation Encoder (Actor \& Critic)}} \\ \hline
Input MLP Layers & 2 \\
Input MLP Hidden Units & 64 \\
\textit{Match Info} MLP Hidden Units & 9 \\
Final FC Hidden Units & 256 \\ \hline
\multicolumn{2}{|l|}{\textbf{Actor Policy Network}} \\ \hline
Main MLP Layers & 4 \\
Main MLP Hidden Units & 256 \\
Positional Encoding Dimension & 256 \\
Pre-Action MLP Layers & 2 \\
Pre-Action MLP Hidden Units & 256 \\
Activation Function & ReLU \\ \hline
\multicolumn{2}{|l|}{\textbf{Critic Network}} \\ \hline
Final Encoder Layers & 2 \\
Final Encoder Hidden Units & 256 \\
Activation Function & ReLU \\ \hline
\multicolumn{2}{|l|}{\textbf{Training Parameters}} \\ \hline
Discount Factor ($\gamma$) & 0.999 \\
GAE Lambda ($\lambda$) & 0.995 \\
PPO Epochs & 2 \\
Num. Mini-batches & 1 \\
Clipping Parameter ($\epsilon$) & 0.2 \\
Value Loss Coefficient & 1.0 \\
Entropy Coefficient & 0.01 \\
Use Orthogonal Initialization & True \\
Initialization Gain & 0.01 \\
Use Feature Normalization & True \\ \hline
\end{tabular}
}
\caption{Configurations and hyperparameters for TiZero's training experiments.}
\label{tab:tizero_training_config}
\end{table}
\begin{table}[H]
\centering
\scalebox{0.8}{
\begin{tabular}{|l|l|}
\hline
\textbf{Attribute} & \textbf{Value} \\ \hline
\multicolumn{2}{|l|}{\textbf{Observation Encoder (Actor \& Critic)}} \\ \hline
Input MLP Layers & 2 \\
Input MLP Hidden Units & 64 \\
\textit{Match Info} MLP Hidden Units & 9 \\
Final FC Hidden Units & 256 \\ \hline
\multicolumn{2}{|l|}{\textbf{Actor Policy Network}} \\ \hline
Main MLP Layers & 4 \\
Main MLP Hidden Units & 256 \\
Positional Encoding Dimension & 256 \\
Pre-Action MLP Layers & 2 \\
Pre-Action MLP Hidden Units & 256 \\
Activation Function & ReLU \\ \hline
\multicolumn{2}{|l|}{\textbf{Critic Network}} \\ \hline
Final Encoder Layers & 2 \\
Final Encoder Hidden Units & 256 \\
Activation Function & ReLU \\ \hline
\multicolumn{2}{|l|}{\textbf{Intrinsic Reward Settings}} \\ \hline
Intrinsic Reward Network Layers & 2 \\
Intrinsic Reward Network Hidden Units & 128 \\
Intrinsic Reward Network Output Dimension & 4 \\
Learning Rate & 0.0005 \\
Clipping Parameter & 0.01 \\ \hline
\multicolumn{2}{|l|}{\textbf{Training Parameters}} \\ \hline
Discount Factor ($\gamma$) & 0.999 \\
GAE Lambda ($\lambda$) & 0.995 \\
PPO Epochs & 2 \\
Num. Mini-batches & 1 \\
Clipping Parameter ($\epsilon$) & 0.2 \\
Value Loss Coefficient & 1.0 \\
Entropy Coefficient & 0.01 \\
Use Orthogonal Initialization & True \\
Initialization Gain & 0.01 \\
Use Feature Normalization & True \\ \hline
\end{tabular}
}
\caption{Configurations and hyperparameters for TiZero-SSIR's training experiments.}
\label{tab:tizero_ssir_training_config}
\end{table}

\end{document}